\documentclass[runningheads]{llncs}
\usepackage{graphicx}
\usepackage{amsfonts}
\usepackage{amssymb}
\usepackage{amsmath}
\usepackage{cases}
\usepackage{array}
\usepackage[section]{placeins}
\usepackage{booktabs}
\usepackage{multirow}
\usepackage{soul}
\usepackage{color}
\usepackage[misc]{ifsym}
\usepackage[colorlinks, anchorcolor=blue, linkcolor=blue, urlcolor=blue, citecolor=blue]{hyperref}
\newcommand*\samethanks[1][\value{footnote}]{\footnotemark[#1]}

\begin{document}
%
\title{Learning Directional Feature Maps for Cardiac MRI Segmentation}
%
%
\author{Feng Cheng \thanks{Equal contribution.} \inst{1} \and
Cheng Chen \samethanks \inst{1} \and
Yukang Wang \inst{1} \and
Heshui Shi \inst{2,3} \and
Yukun Cao \inst{2,3} \and
Dandan Tu \inst{4} \and
Changzheng Zhang\inst{4} \and
Yongchao Xu\inst{1}}
\authorrunning{F. Cheng et al.}

%
\institute{School of Electronic Information and Communications, Huazhong University of Science and Technology, Wuhan, China\and
Department of Radiology, Union Hospital, Tongji Medical College, Huazhong University of Science and Technology, Wuhan, China \and
Hubei Province Key Laboratory of Molecular Imaging, Wuhan, China \and
HUST-HW Joint Innovation Lab, Wuhan, China \\
\email{yongchaoxu@hust.edu.cn}}


%
\maketitle              
\begin{abstract}
Cardiac MRI segmentation plays a crucial role in clinical diagnosis
for evaluating personalized cardiac performance parameters. Due to the
indistinct boundaries and heterogeneous intensity distributions in the
cardiac MRI, most existing methods still suffer from two aspects of
challenges: inter-class indistinction and intra-class
inconsistency. To tackle these two problems, we propose a novel method
to exploit the directional feature maps, which can simultaneously
strengthen the differences between classes and the similarities within
classes.  Specifically, we perform cardiac segmentation and learn a
direction field pointing away from the nearest cardiac tissue boundary
to each pixel via a direction field (DF) module.
Based on the learned direction field, we then propose a feature
rectification and fusion (FRF) module to improve the original
segmentation features, and obtain the
final segmentation. The proposed modules are simple yet effective and
can be flexibly added to any existing segmentation network without
excessively increasing time and space complexity. We evaluate the
proposed method on the 2017 MICCAI Automated Cardiac Diagnosis
Challenge (ACDC) dataset and a large-scale self-collected dataset,
showing good segmentation performance and robust generalization
ability of the proposed method. The code is publicly available at \url{https://github.com/c-feng/DirectionalFeature}.
\keywords{Cardiac Segmentation
  \and Deep Learning \and Direction Field.}
\end{abstract}
\section{Introduction}

Cardiac cine Magnetic Resonance Imaging (MRI) segmentation is of great
importance in disease diagnosis and surgical planning. Given the
segmentation results, doctors can obtain the cardiac diagnostic
indices such as myocardial mass and thickness, ejection fraction and
ventricle volumes more efficiently. Indeed, manual segmentation is the
gold standard approach. However, it is not only time-consuming but
also suffers from the inter-observer variations. Hence, the automatic
cardiac cine MRI segmentation is desirable in the clinic.


\par In the past decade, deep convolutional neural networks (CNNs)
based methods have achieved great successes in both natural and
medical image segmentation.  U-Net~\cite{ronneberger2015u} is one of
the most successful and influential method in medical image
segmentation.  Recent works typically leverage the U-shape networks
and can be roughly divided into 2D and 3D methods.
2D methods take a single 2D slice as input while 3D methods utilize
the entire volume.
nnU-Net~\cite{isensee2019nnu} adopts the model fusion strategy of 2D
U-Net and 3D U-Net, which achieves the current state-of-the-art
performance in cardiac segmentation.  However, the applicability is
somewhat limited since it requires a high cost of memory and
computation.
\par
\begin{figure*}[t]
\centering
\includegraphics[width=0.86\linewidth]{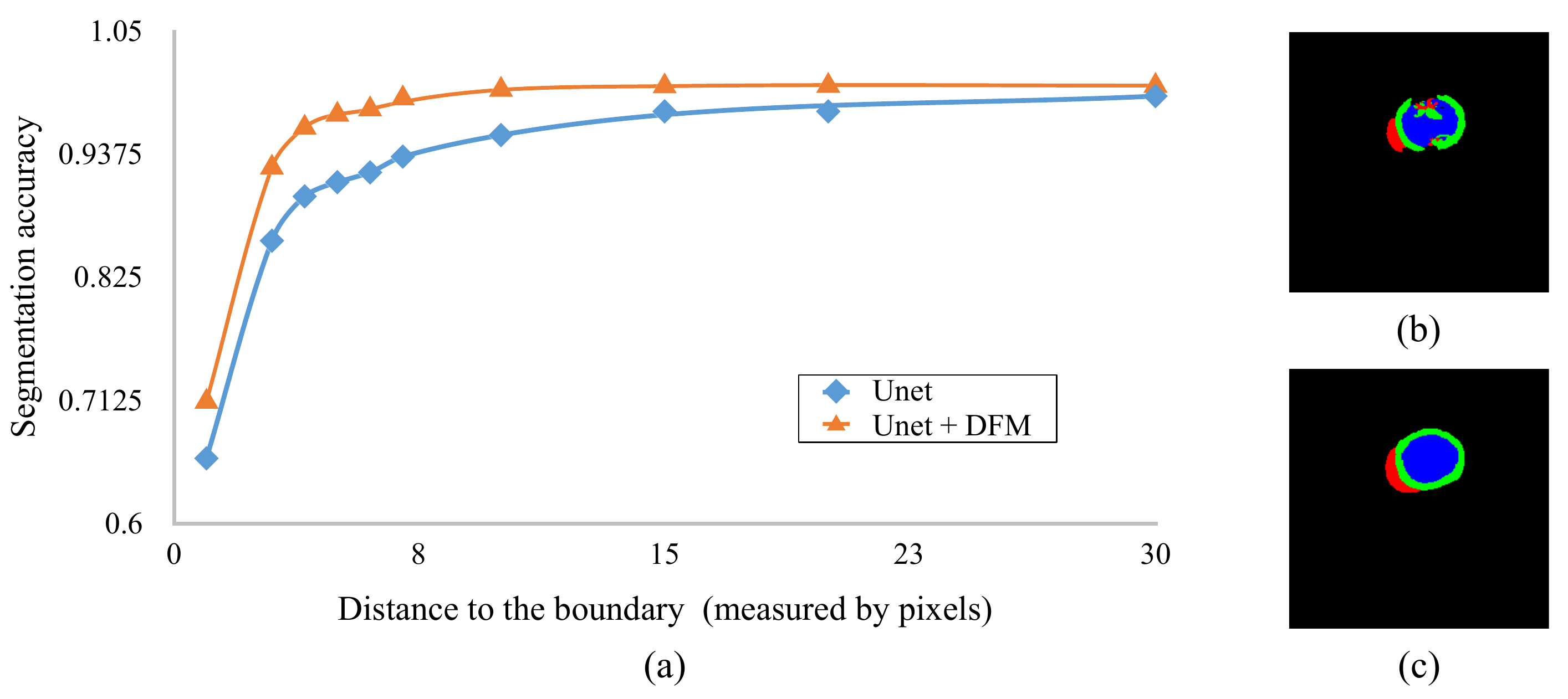}
\caption{(a) is the comparison of segmentation accuracy between U-Net
  and the proposed method at different distances from pixel to
  boundary; (b) and (c) are the segmentation visualizations of U-Net
  and the proposed method, respectively. Compared with the original
  U-Net, the proposed method effectively mitigate the problems of
  inter-class indistinction and intra-class inconsistency.}
\label{fig:distance-acc}
\end{figure*}

 The MRI artifacts such as intensity inhomogeneity and fuzziness
may make it indistinguishable between pixels near the boundary,
leading to the problem of inter-class indistinction. As depicted in
Fig.~\ref{fig:distance-acc}(a), we observe that the cardiac MRI
segmentation accuracy drops dramatically for those pixels close to the
boundary.
Meanwhile, due to the lack of restriction on the spatial relationship
between pixels, the segmentation model may produce some anatomical
implausible errors (see Fig.~\ref{fig:distance-acc}(b) for an
example).  In this paper, we propose a novel method to improve the
segmentation feature maps with directional information, which can
significantly improve
the inter-class indistinction as well as cope with the intra-class
inconsistency. Extensive experiments demonstrate that the proposed method
achieves good performance and is robust under cross-dataset validation.

Recent approaches in semantic segmentation have been devoted to
handling the inter-class indistinction and the intra-class
inconsistency.
Ke \textit{et al.}~\cite{ke2018adaptive} define the concept of
adaptive affinity fields (AAF) to capture and match the semantic
relations between neighboring pixels in the label space. Cheng \textit{et al.}~\cite{cheng2020eccv} explore the boundary and segmentation mask information to improve the inter-class indistinction problem for instance segmentation.
Shusil
\textit{et al.}~\cite{dangi2019distance} propose a multi-task learning
framework to perform segmentation along with a pixel-wise distance map
regression. This regularization method takes the distance from the
pixel to the boundary as auxiliary information to handle the problem
of inter-class indistinction.  Nathan \textit{et
  al.}~\cite{painchaud2019cardiac} propose an adversarial variational
auto-encoder to assure anatomically plausible, whose latent space
encodes a smooth manifold on which lies a large spectrum of valid
cardiac shapes, thereby indirectly solving the intra-class
inconsistency problem.

\par Directional information has been recently explored in different
vision tasks. For instance, TextField~\cite{xu2019textfield} and
DeepFlux~\cite{wang2019deepflux} learn similar direction fields on
text areas and skeleton context for scene text detection and skeleton
extraction, respectively. They directly construct text instances or
recover skeletons from the direction field. However, medical images
are inherently different from natural images. Such segmentation
results obtained directly from the direction field are not accurate
for the MRI segmentation task.
In this paper, we propose to improve the original segmentation
features guided by the directional information for better cardiac MRI
segmentation.




%
%
%
\begin{figure*}[t]
\centering
\includegraphics[width=1.0\linewidth]{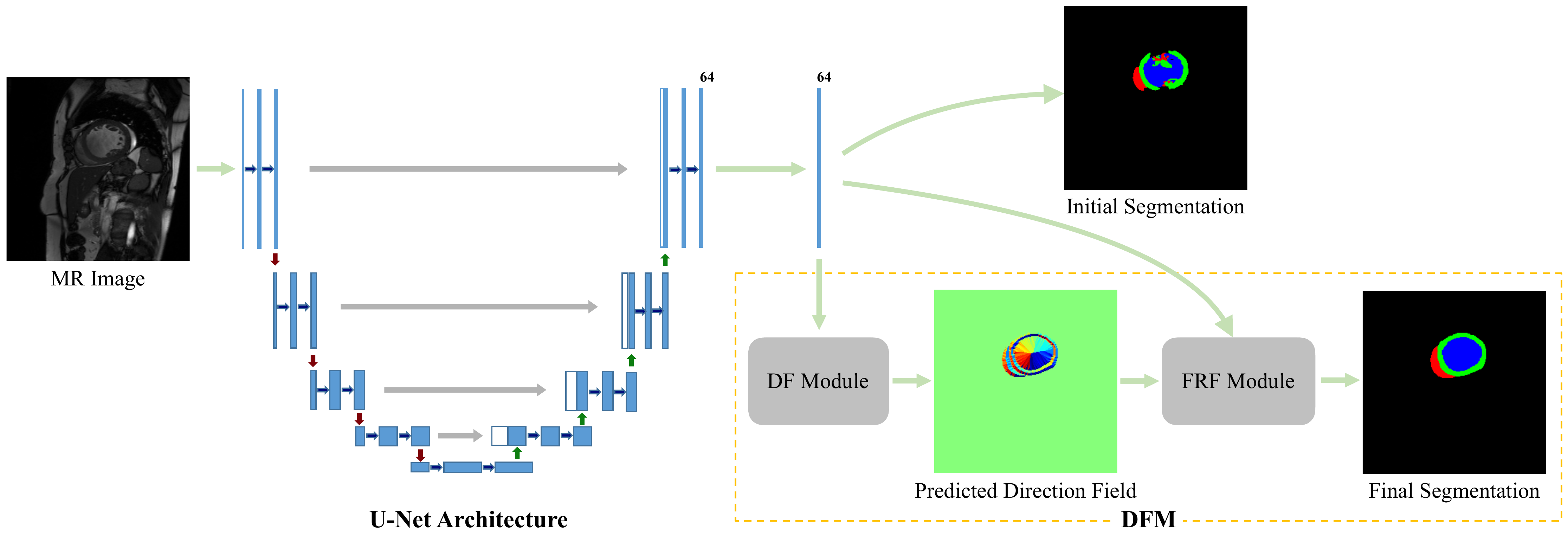}
\caption{Pipeline of the proposed method. Given an image, the
  network predicts an initial segmentation map from U-Net and a
  direction field (DF) (visualized by its direction information),
  based on which  we reconstruct and fuse the original segmentation
  features via a feature rectification and fusion (FRF) module to
  produce the final segmentation. }
\label{fig:pipeline}
\end{figure*}

\section{Method}

Inter-class indistinction and intra-class inconsistency are commonly
found in both natural and medical image segmentation.
Meanwhile, segmentation models usually learn individual
representations and thus lack of restrictions on the relationship
between pixels.  We propose a simple yet effective method to exploit
the directional relationship between pixels, which can simultaneously
strengthen the differences between classes and the similarities within
classes. The pipeline of the proposed method, termed as
DFM, is depicted in Fig.~\ref{fig:pipeline}. We adopt
U-Net~\cite{ronneberger2015u} as our base segmentation framework.
Given an input image, the network produces the initial segmentation
map. Meanwhile, we apply a direction field (DF) module to learn the
direction field with the shared features from U-Net. A feature
rectification feature (FRF) module is then proposed to combine the
initial segmentation feature with the learned direction field to
generate the final improved segmentation result.

\subsection{DF Module to Learn a Direction Field}
\begin{figure*}[t]
\centering
\includegraphics[width=1.0\linewidth]{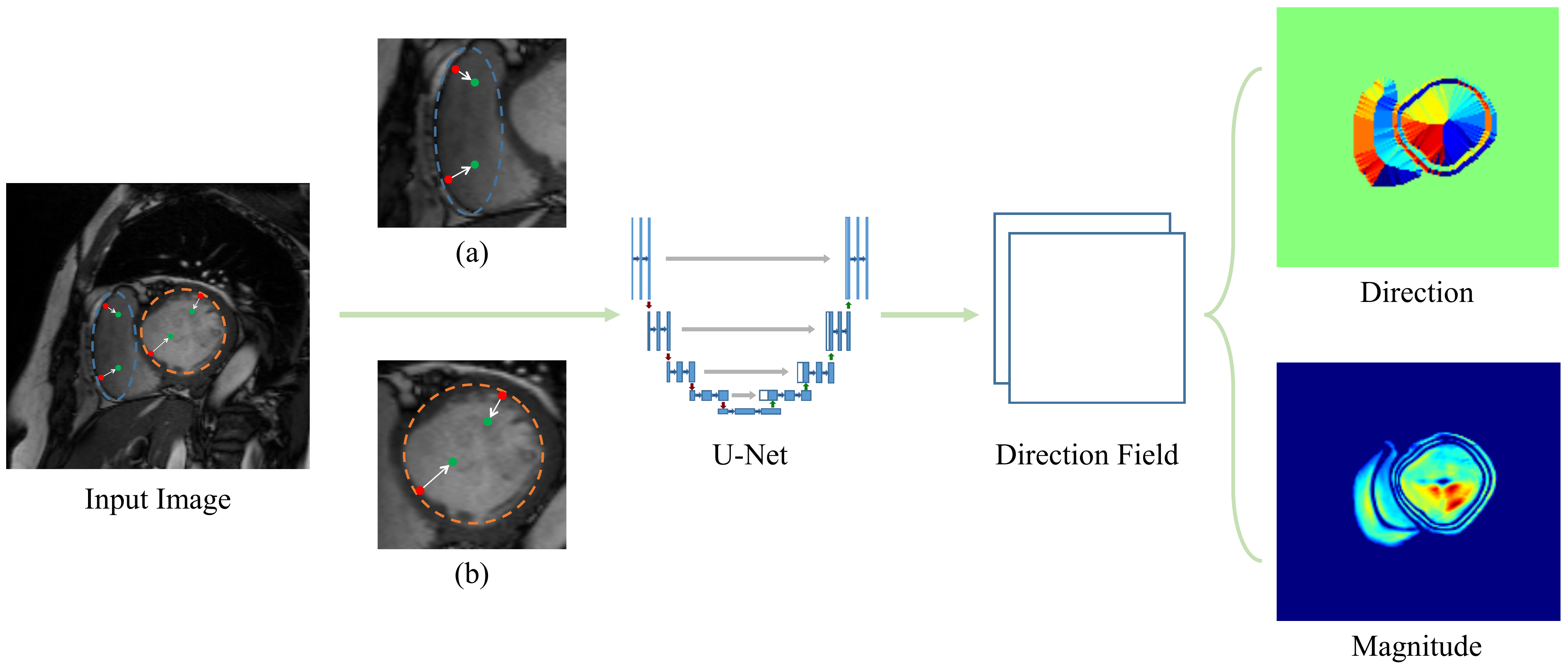}
\caption{Illustration of the DF module. Given an image, the
  network predicts a novel direction field in terms of an image of
  two-dimensional vectors. (a) and (b) show the vector from the
  nearest boundary pixel to the current pixel. We calculate and
  visualize the direction and magnitude information of the direction
  field on the right side.}
\label{fig:DF-module}
\end{figure*}

We first detail the notation of the direction field.  As shown in
Fig.~\ref{fig:DF-module}(a-b), for each foreground pixel $p$,
we find its nearest pixel $b$ lying on the cardiac tissue
boundary and then normalize the direction vector $\overrightarrow{bp}$ pointing from $b$ to $p$ by the distance between $b$ and $p$.
We set the background pixels to $(0,0)$. Formally, the direction field $DF$ for each pixel $p$ in the image domain $\Omega$ is given by:
\begin{equation}
    DF(p) =
    \begin{cases}
        \frac{\overrightarrow{bp}}{|\overrightarrow{bp}|} & p \in foreground, \\
        (0,0) & otherwise.
    \end{cases}
\end{equation}

We propose a simple yet effective DF module to learn the above
direction field, which is depicted in Fig.~\ref{fig:DF-module}. This
module is made up of a $1 \times 1$ convolution, whose input is the
64-channel feature extracted by U-Net and output is the two-channel
direction field.
It is noteworthy that we can obtain the ground truth of the direction field from the annotation easily by distance transform algorithm.

\subsection{FRF Module for Feature Rectification and Fusion}

\begin{figure*}[t]
\centering
\includegraphics[width=0.95\linewidth]{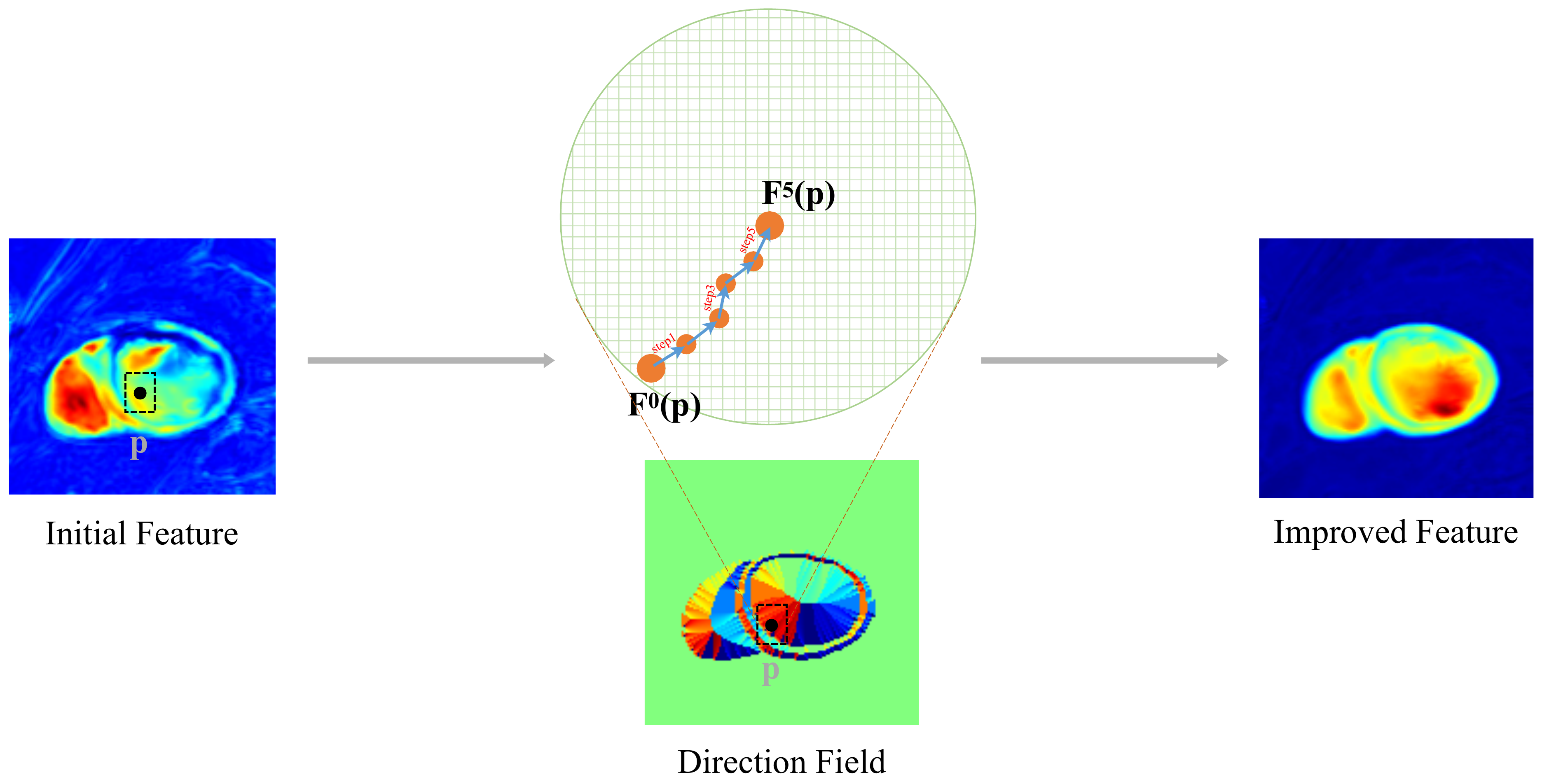}
\caption{Schematic illustration of the FRF module. We use the learned
  direction field to guide the feature rectification. The feature for
  pixels close to the boundary is rectified with the feature (having the
  same semantic category) far from the boundary.}
\label{fig:heatmap}
\end{figure*}

The direction field predicted by the DF module reveals the directional
relationship between pixels and provides a unique direction vector
that points from the boundary to the central area for each
pixel. Guided by these direction vectors, we propose a Feature
Rectification and Fusion (FRF) module to utilize the characteristics
of the central area to rectify the errors in the initial segmentation
feature maps step by step.  As illustrated in Fig.~\ref{fig:heatmap},
the N-steps improved feature maps $F^N \in \mathbb{R}^{C \times H
  \times W}$ are obtained with the initial feature maps $F^0 \in
\mathbb{R}^{C \times H \times W}$ and the predicted direction field
$DF \in \mathbb{R}^{2 \times H \times W}$ step by step.
Concretely, the improved feature of the pixel $p$ is updated
iteratively by the feature of the position that $DF(p)$ points to,
which is calculated by the bilinear interpolation. In other words,
$F(p)$ is rectified by the features of the central area gradually.
The whole procedure is formalized as below:
\begin{equation}
    \forall p \in \Omega, \, F^{k}(p) \, = \, F^{(k-1)}\big(p_x + DF(p)_x, p_y + DF(p)_y\big), 
\label{eq:frf}
\end{equation}
where $1 \leq k \leq N$ denotes the current step, $N$ is the total steps (set to 5 if not stated otherwise), and $p_x$ (\textit{resp.} $p_y$) represents the $x$ (\textit{resp.} $y$) coordinate of the pixel $p$.

After performing the above rectification process, we concatenate $F^N$ with $F^0$, and then apply the final classifier on the concatenated feature maps to predict the final cardiac segmentation.

\subsection{Training Objective}
The proposed method involves loss function on the initial segmentation $L_{CE}^i$, final segmentation $L_{CE}^f$, and direction field $L_{DF}$.
We adopt the general cross-entropy $L_{CE}$ as the segmentation loss to
encourage class-seperate feature, which is commonly used in semantic segmentation.
Formally, $L_{CE}$ is given by
    $L_{CE} = -\sum_{i}{y_i log(\hat{y}_i)}$,
{where $y_i$ and $\hat{y}_i$ denote the ground truth and the prediction}, respectively.
For the loss to supervise the direction field learning,
we choose the $L_2$-norm
distance and angle distance as the training objective:
\begin{equation}
    L_{DF} = \sum_{p \in \Omega}w(p)(||DF(p) - \hat{DF}(p)||_2 + \alpha \times ||cos^{-1}\langle DF(p),\hat{DF}(p)\rangle||^2)
\end{equation}
{where $\hat{DF}$ and $DF$ denote the predicted direction field and the corresponding ground truth}, respectively, $\alpha$ is a hyperparameter to balance the $L_2$-norm distance and angle distance, and is to 1 in all experiments, and $w(p)$ represents the weight on pixel $p$, which is calculated by:
\begin{equation}
    w(p) =
    \begin{cases}
        \frac{\sum_{i=1}^{N_{cls}}{|C_i|}}{N_{cls} \cdot |C_i|} & p \in C_i, \\
        1 & otherwise,
    \end{cases}
\end{equation}
where $|C_i|$ denotes the total number of pixels with label $i$ and $N_{cls}$ is the number of classes. The
overall loss $L$ combines $L_{CE}$ and $L_{DF}$ with a balance factor $\lambda=1$:
\begin{equation}
    L = L_{CE}^i + L_{CE}^f + \lambda L_{DF}
\label{eq:overallloss}
\end{equation}

\section{Experiments}
\subsection{Datasets and Evaluation Metrics}
\label{subsec:dataset}
\par \noindent \textbf{Automatic Cardiac Diagnosis Challenge (ACDC)
  Dataset} contains cine-MR images of 150 patients, split into 100
train images and 50 test images. These patients are divided into 5
evenly distributed subgroups: normal, myocardial infarction, dilated
cardiomyopathy, hypertrophic cardiomyopathy and abnormal right
ventricle, available as a part of the STACOM 2017 ACDC challenge.  The
annotations for the 50 test images are hold by the challenge
organizer. We also further divide the 100 training images into 80\%
training and 20\% validation with five non-overlaping folds to perform
extensive experiments.

\par \noindent \textbf{Self-collected Dataset} consists of
more than 100k 2D images that we collected from 531 patient cases. {All the data was labeled by a team of medical experts.} The patients
are also divided into the same 5 subgroups as ACDC.
A series of short axis slices cover LV from the base to the apex, with a thickness of 6 mm and a flip angle of 80$^{\circ}$. The magnitude field strength of images is 1.5T and the spatial resolution is 1.328 $mm^{2}$/pixel. We also split all the patients into 80\% training and 20\% test.

\par \noindent \textbf{Evaluation metrics:} We adopt the widely used
3D Dice coefficient and Hausdorff distance to benchmark the proposed
method.

\subsection{Implementation Details}
The network is trained by minimizing the proposed loss function in
Eq.~\eqref{eq:overallloss} using ADAM optimizer~\cite{kingma2014adam}
with the learning rate set to 10$^{-3}$. The network weights are
initialized with~\cite{he2015delving} and
trained for 200 epochs. Data augmentation is applied to prevent
over-fitting including: 1) random translation with the maximum
absolute fraction for horizontal and vertical translations both set to
0.125; 2) random rotation with the random angle between
-180$^{\circ}$ and 180$^{\circ}$. The batch size is set to 32 with the
resized $256\times256$ inputs.

\begin{table}[t]
\caption{Performance on ACDC dataset (train/val split) and Self-collected dataset.}\label{acdc&&union}
\begin{tabular*}{\hsize}{@{}@{\extracolsep{\fill}}c|c|cccc|ccccc@{}}
\toprule[1pt]
\multirow{2}{*}{}  & \multirow{2}{*}{Methods}  & \multicolumn{4}{c|}{Dice Coefficient}  & \multicolumn{4}{c}{Hausdorff Distance (mm)}  \\ \cline{3-10}
                                                                                    &   & LV      & RV      & MYO     & Mean      & LV      & RV      & MYO     & Mean      \\ \toprule[1pt]
\multirow{2}{*}{ACDC Dataset}                                                       & U-Net      & 0.931 & 0.856 & 0.872 & 0.886 & 24.609  & 30.006 & 14.416 & 23.009 \\ \cline{2-10}
                                                                                    &\textbf{Ours} & \textbf{0.949} & \textbf{0.888} & \textbf{0.911} & \textbf{0.916} & \textbf{3.761} & \textbf{6.037} & \textbf{10.282} & \textbf{6.693}    \\ \hline
\multirow{2}{*}{\begin{tabular}[c]{@{}c@{}}Self-Collected\\   Dataset\end{tabular}} & U-Net & 0.948  & 0.854 & 0.906  & 0.903   & 2.823 & 3.691  & 2.951 & 3.155   \\ \cline{2-10}
                                                                                    & \textbf{Ours} & \textbf{0.949} & \textbf{0.859} & \textbf{0.909} & \textbf{0.906} & \textbf{2.814} & \textbf{3.409} & \textbf{2.683} & \textbf{2.957}      \\
\bottomrule[1pt]
\end{tabular*}

\end{table}

\begin{table}[t]
\caption{Results on the ACDC leaderboard (sorted by Mean Hausdorff Distance).}\label{leadboard}
\begin{tabular*}{\hsize}{@{}@{\extracolsep{\fill}}cccc@{}}
\toprule[1pt]
Rank       & User                  & Mean DICE     & Mean HD (mm)       \\ \hline\hline
1          & Fabian Isensee~\cite{isensee2019nnu}       & 0.927         & 7.8           \\ \hline
2          & Clement Zotti~\cite{zotti2018convolutional}         & 0.9138        & 9.51          \\ \hline
\textbf{3} & \textbf{Ours}         & \textbf{0.911} & \textbf{9.92} \\ \hline
4          & Nathan Painchaud~\cite{painchaud2019cardiac}    & 0.911         & 9.93          \\ \hline
5          & Christian Baumgartner~\cite{baumgartner2017exploration} & 0.9046        & 10.4          \\ \hline
6          & Jelmer Wolterink~\cite{wolterink2017automatic}      & 0.908         & 10.7          \\ \hline
7          & Mahendra Khened~\cite{khened2019fully}       & 0.9136        & 11.23         \\ \hline
8          & Shubham Jain~\cite{patravali20172d}          & 0.8915        & 12.07         \\
\bottomrule[1pt]
\end{tabular*}
\end{table}

\begin{table}[t]
\caption{Comparisons with different methods aiming to alleviate the inter-class indistinction and intra-class inconsistency on ACDC dataset (train/val split).}\label{AAF-distancemap}
\begin{tabular*}{\hsize}{@{}@{\extracolsep{\fill}}c|cccc|cccc@{}}
\toprule[1pt]
\multirow{2}{*}{Methods}
                  & \multicolumn{4}{c|}{Dice Coefficient}                                               & \multicolumn{4}{c}{Hausdorff Distance (mm)}                            \\ \cline{2-9}
          & LV              & MYO             & RV              & Mean              & LV            & MYO            & RV            & Mean              \\ \hline
U-Net              & 0.931          & 0.856          & 0.872          & 0.886          & 24.609         & 30.006         & 14.416          & 23.009         \\

AAF~\cite{ke2018adaptive}               & 0.928          & 0.853          & 0.891          & 0.891          & 13.306         & 14.255          & 13.969          & 13.844          \\

DMR~\cite{dangi2019distance}      & 0.937          & 0.880            & 0.892          & 0.903             & 7.520          & 9.870           & 12.385        & 9.925             \\ \hline

\textbf{U-Net+DFM} & \textbf{0.949} & \textbf{0.888} & \textbf{0.911} & \textbf{0.916} & \textbf{3.761} & \textbf{6.037} & \textbf{10.282} & \textbf{6.693} \\
\bottomrule[1pt]
\end{tabular*}
\end{table}

\begin{figure*}[t]
\centering
\includegraphics[width=1.0\linewidth]{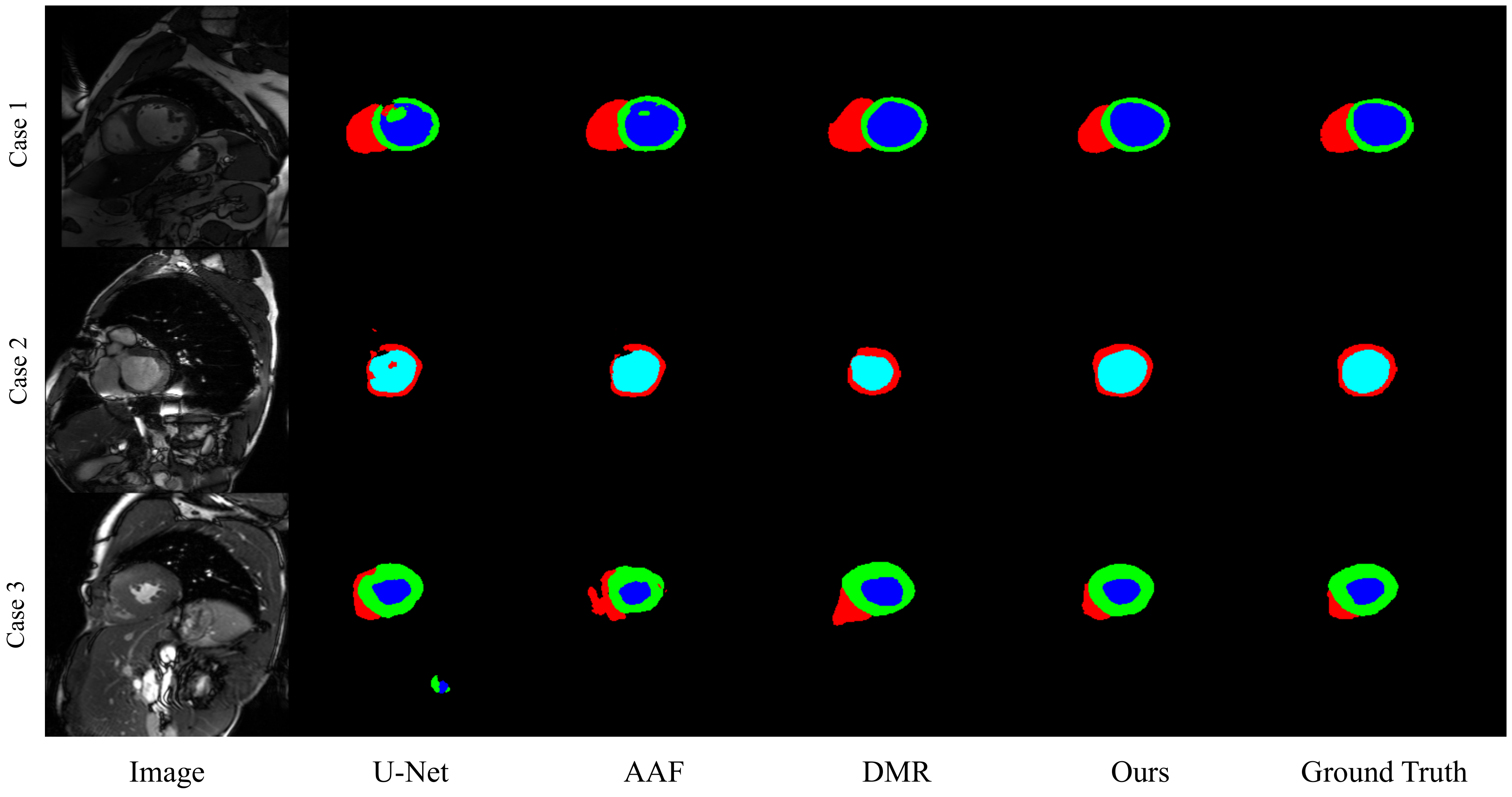}
\caption{Qualitative comparison on ACDC dataset. The proposed DFM achieves more accurate results along with better smoothness and continuity in shape.}
\label{fig:comparison}
\end{figure*}

\subsection{In-dataset Results}

\par We first evaluate the proposed method on the ACDC dataset
(train/val split described in Sec.~\ref{subsec:dataset}) and the
self-collected dataset. Tab.~\ref{acdc&&union} presents the
performance of the proposed DFM on two datasets. LV, RV and MYO
represent the left ventricle, right ventricle and myocardium,
respectively. Our approach consistently improves the baseline (U-Net),
demonstrating its effectiveness. To further make comparison with the
current state-of-the-arts methods, we submit the results to the ACDC
leadboard. As shown in Tab.~\ref{leadboard}, compared with those
methods that rely on well-designed networks (\textit{e.g.} 3D network
in~\cite{isensee2019nnu} and Grid-like CNN in~\cite{zotti2018convolutional}) or multi-model fusion, the proposed DFM
achieves competitive performance with only two simple yet effective
modules added to the baseline U-Net.  We also compare our approach
with other methods dedicated to alleviate the inter-class
indistinction and intra-class inconsistency. As depicted in
Tab.~\ref{AAF-distancemap}, the proposed DFM significantly outperforms
these methods. Some qualitative comparisons are given in
Fig.~\ref{fig:comparison}.

\begin{table}[t]
\caption{Cross-dataset evaluation compared with the original U-Net.}
\label{cross_test}
\begin{tabular*}{\hsize}{@{}@{\extracolsep{\fill}}c|cc|cc@{}}
\toprule[1pt]
\multirow{2}{*}{Methods} & \multicolumn{2}{c|}{ACDC to self-collected dataset} & \multicolumn{2}{c}{Self-collected dataset to ACDC} \\ \cline{2-5}
                         & Mean Dice                    & Mean HD (mm)                  & Mean Dice                    & Mean HD (mm)                  \\ \hline
U-Net                    & 0.832                        & 25.553                       & 0.803                        & 4.896                        \\ \hline
\textbf{U-Net+DFM}       & \textbf{0.841}               & \textbf{17.870}              & \textbf{0.820}               & \textbf{4.453}               \\ \bottomrule[1pt]
\end{tabular*}
\end{table}

\begin{table}[!h]
\caption{Ablation study on the number of steps $N$ on ACDC dataset.}\label{step}
\begin{tabular*}{\hsize}{@{}@{\extracolsep{\fill}}ccccccc@{}}
\toprule[1pt]
\multicolumn{2}{c}{Number of steps}      & 0    & 1    & 3    & \textbf{5}    & 7    \\ \hline
\multicolumn{2}{c}{Mean Dice} & 0.910 & 0.913 & 0.914   & \textbf{0.916} & 0.910 \\ \hline
\multicolumn{2}{c}{Mean HD (mm)}   & 10.498 & 17.846 & 9.026 & \textbf{6.693}    & 13.880   \\
\bottomrule[1pt]
\end{tabular*}
\label{tab:ablation}
\end{table}

\subsection{Cross Dataset Evaluation and Ablation Study}
To analyze the generalization ability of the proposed DFM, we
performed a cross-dataset segmentation evaluation.
Results listed in Tab.~\ref{cross_test} show that the proposed DFM can
consistently improve the cross-dataset performance compared with the
original U-Net, validating its generalization ability and robustness.
We also conduct ablation study on ACDC dataset to explore how the
number of steps $N$ involved in the FRF module influences the performance. As
shown in Tab.~\ref{tab:ablation}, the setting of $N=5$ gives the best
performance.

\section{Conclusion}
In this paper, we explore the importance of directional information
and present a simple yet effective method for cardiac MRI
segmentation. We propose to learn a direction field, which
characterizes the directional relationship between pixels and
implicitly restricts the shape of the segmentation result. Guided by
the directional information, we improve the segmentation feature maps
and thus achieve better segmentation accuracy. Experimental results
demonstrate the effectiveness and the robust generalization ability of
the proposed method.

\section*{Acknowledgement}
This work was supported in part by the Major Project for New Generation of AI under Grant no. 2018AAA0100400,
NSFC 61703171, and NSF of Hubei Province of China
under Grant 2018CFB199.
Dr. Yongchao Xu was supported by
the Young Elite Scientists Sponsorship Program by CAST.


\end{document}